%% file: arxiv.tex
\documentclass[conference]{IEEEtran}
\usepackage{rss}
\usepackage{makecell}
\usepackage{svg}
\usepackage{algorithm}
\usepackage{algpseudocode}
\usepackage{fontawesome5}
\usepackage{float}

\definecolor{somegray}{rgb}{0.5, 0.5, 0.5}
\newcommand{\darkgrayed}[1]{\textcolor{somegray}{#1}}
\makeatletter
\newcommand*\titleheader[1]{\gdef\@titleheader{#1}}
\AtBeginDocument{%
  \let\st@red@title\@title
  \def\@title{%
    \vskip-2.2em
    \bgroup\normalfont\large\centering\@titleheader\par\egroup
    \vskip1.0em\st@red@title}
}
\makeatother
\title{Learning Agile Quadrotor Flight in the Real World}

\titleheader{\darkgrayed{This paper has been accepted for publication at Robotics: Science and Systems, 2026}}
\begin{document}
\input{figures/eyecatcher}

\author{Yunfan Ren, Zhiyuan Zhu, Jiaxu Xing, Davide Scaramuzza\\
    Robotics and Perception Group, University of Zurich, Switzerland}

\maketitle
\input{sections/abstract}

\IEEEpeerreviewmaketitle
\section*{Supplementary Materials}
\noindent\faGlobe~\textbf{Website:} \url{https://rpg.ifi.uzh.ch/lafr/} \\
\faGithub~\textbf{Code:} \url{https://github.com/uzh-rpg/LAFR}

\input{sections/introduction}
\input{sections/related_work}
\input{sections/methodology}
\input{sections/experiments}

\input{sections/discussion}
\input{sections/acknowlegement}

\bibliographystyle{ieeetr}
\bibliography{references}
\clearpage
\end{document}

%% file: figures/eyecatcher.tex
\makeatletter
\g@addto@macro\@maketitle{
  \captionsetup{type=figure}\setcounter{figure}{0}
  \def\mycolspace{1.9mm}
  \centering
  \vspace{1em}
  \includegraphics[width=\linewidth]{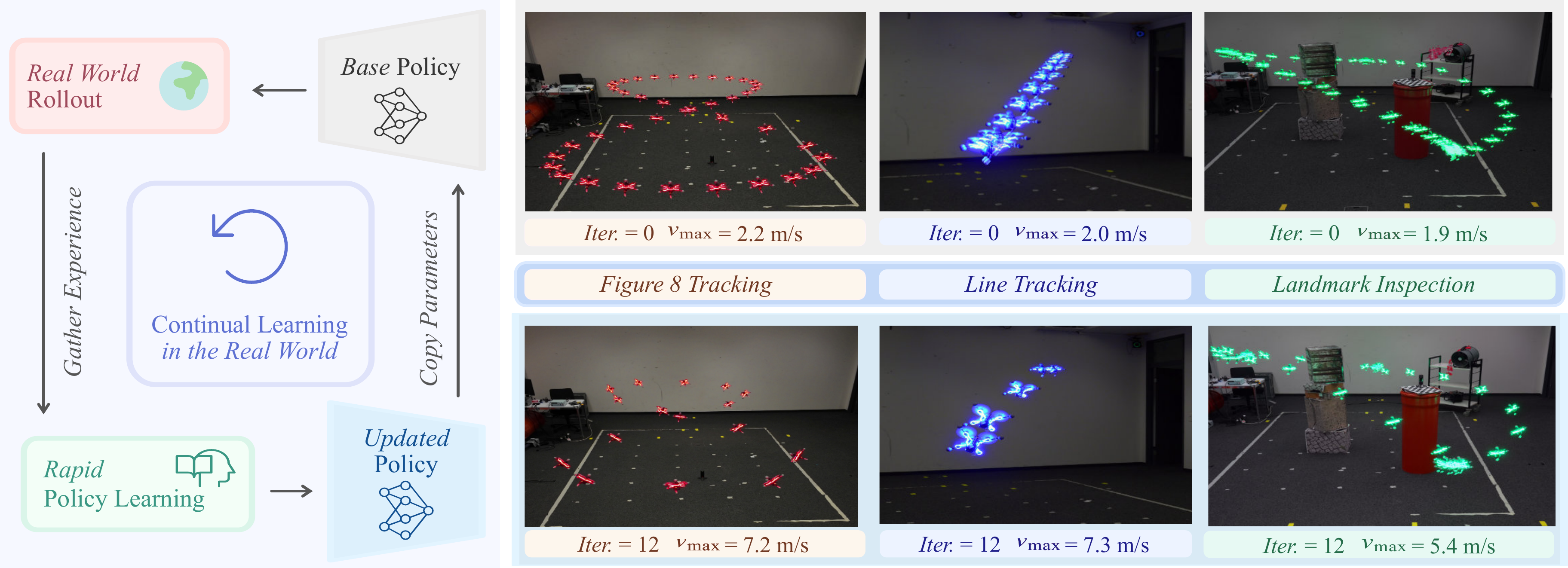}
  \caption{\textbf{Autonomous online evolution of agile flight.}
    The figure illustrates the continuous loop between \textit{Real-World Rollouts} and \textit{Rapid Policy Learning}, comparing the initial conservative flights (Iter 0, top) with the evolved agile behaviors (Iter 12, bottom).
    Trajectories are composited with frames sampled at uniform \SI{0.5}{\second} intervals. Consequently, trajectory density reflects flight speed, where sparser segments indicate higher velocity.
    Driven by this online adaptation, the system pushes its physical limits without human intervention, tripling its speed within \textbf{\SI{100}{\second}} of flight time across three tasks.
  }
  \label{fig:catcheye}
  \vspace{-0.4cm}
}
\makeatother

%% file: sections/abstract.tex
\begin{abstract}
    Learning-based controllers have achieved impressive performance in agile quadrotor flight but typically rely on large-scale simulation training, necessitating accurate system identification for effective sim-to-real transfer.
    However, even with precise modeling, fixed policies remain susceptible to out-of-distribution scenarios, ranging from external aerodynamic disturbances to internal hardware degradation. To ensure safety under these evolving uncertainties, such controllers are forced to operate with conservative safety margins, inherently constraining their agility outside of controlled settings.
    While online adaptation offers a potential remedy, safely pushing the platform toward its agility envelope remains a critical bottleneck due to data scarcity and safety risks.
    To bridge this gap, we propose a self-adaptive framework that eliminates the need for precise system identification or offline sim-to-real transfer.
    We introduce Adaptive Temporal Scaling (ATS) to actively push the platform toward the agility ceiling exposed by its control interface, and employ online residual learning to augment a simple nominal model. Based on the learned hybrid model, we further propose Real-World Anchored Short-Horizon Backpropagation Through Time (RASH-BPTT) to achieve efficient and robust in-flight policy updates.
    Extensive experiments demonstrate that our quadrotor reliably executes agile maneuvers near the saturation limit of the CTBR command interface. The system evolves a conservative base policy from a peak speed of \SIrange{2.0}{7.3}{\meter\per\second} within approximately \SI{100}{\second} of flight time. These findings underscore that real-world adaptation serves not merely to compensate for modeling errors, but as a practical mechanism for sustained performance improvement in aggressive flight regimes.

\end{abstract}

%% file: sections/introduction.tex
\section{Introduction}

The domain of agile control has witnessed a paradigm shift with the advent of data-driven policies, which now rival or exceed the capabilities of classical model-based methods~\cite{tang2025deep, loquercio2021learning, kaufmann23champion, lee2020learning, miki2022learning, he2024agile}.
Inspired by human skill acquisition, most existing methods rely on imitation learning from expert demonstrations~\cite{chi2025diffusion, chi2024universal, xing2024contrastive} or reinforcement learning through large-scale trial-and-error~\cite{kaufmann23champion, miki2022learning, he2024agile}.
In practice, these policies are typically trained in simulation with extensive domain randomization~\cite{tobin2017domain} to approximate real-world variability and are kept fixed during real-world deployment.
As a consequence, once deployed, the controller relies entirely on prior training experience rather than learning and adapting in the real world.
More broadly, learning directly through real-world experience has begun to emerge at the task-planning level~\cite{qu2026pragmatist}. However, the capability to continually improve low-level control through real-world interaction, which is central to human motor learning, remains largely underexplored.

This limitation becomes especially critical in the context of agile quadrotor flight.
Achieving high-speed, aggressive maneuvers requires precise control under tightly coupled and highly nonlinear dynamics, leaving little margin for {model mismatch or unknown external} disturbances~\cite{hanover2024autonomous}.

However, real-world flight conditions are inherently non-stationary. Factors such as hardware wear, battery depletion, and complex aerodynamic effects introduce inevitable discrepancies between the training model and physical reality that fixed policies cannot mitigate.
Consequently, relying on a {fixed policy} directly degrades performance and reliability, often preventing the system from safely reaching its agility potential. In contrast, enabling controllers to learn directly in the real world fundamentally expands the utility of these systems.
This capability empowers the agent to adapt not only to external environmental shifts but also to internal hardware variations, continuously exploiting platform-specific {properties to push performance toward the agility ceiling imposed by the control interface without requiring expensive offline system identification~\cite{o2022neural}.}

However, learning directly in the real world introduces dual challenges of safety and data scarcity. {First, agile quadrotors operate near the limits of stability, where even minor deviations in state or control can trigger catastrophic failure. Second, learning in the physical world demands high sample efficiency because data collection is costly and constrained by limited battery life, permitting only a few interactions. Consequently, a safe, adaptive, and efficient approach is required.}

Existing work has made significant progress toward learning real-world agile quadrotor control.
Sim-to-real approaches~\cite{aljalbout2025reality} based on extensive domain randomization~\cite{tobin2017domain} expose policies to diverse dynamics and disturbances during training, enabling them to tolerate moderate modeling errors.
While effective in controlled settings, their performance can degrade significantly when real-world conditions deviate from the randomized training distribution.
To further reduce the sim-to-real gap, Real2Sim2Real pipelines refine the simulation model using real-world data and retrain the policy offline before deployment~\cite{lee2020learning, bauersfeld2021neurobem, pmlr-v305-sobanbabu25a}.
Although effective, these methods require substantial data collection and repeated retraining cycles, which limit their practicality when system dynamics evolve or rapid adaptation is required.
    {Recent advancements have begun to leverage differentiable simulation for rapid real-world adaptation of quadrotor control policies, notably in~\cite{pan2026learning}. However, while this method facilitates rapid refinement during deployment, it prioritizes disturbance rejection over pushing the platform toward its agility envelope. Consequently, the resulting performance remains conservative, typically around \SIrange{1}{2}{\meter\per\second}, making the policy unsuitable for agile, high-performance behaviors.}

\subsection*{Contributions}
In this work, we propose a self-adaptive framework that bridges the gap between robustness and performance, enabling continuous, on-policy improvement directly in the real world without the need for precise system identification.
Our approach relies on two key mechanisms to address the challenges of safe exploration and efficient learning.
First, to tackle the safety-exploration paradox, we introduce \textit{Adaptive Temporal Scaling (ATS)}.
Unlike fixed-task baselines, ATS actively incentivizes the quadrotor to explore its physical capability boundaries by dynamically adjusting the aggressiveness of the reference trajectory based on the agent's current competence.
Second, to ensure efficient learning, we employ a hybrid dynamics model that augments a simple nominal model with {online residual learning}.
{By introducing \textit{Real-World Anchored Short-Horizon Backpropagation Through Time (RASH-BPTT)} using these hybrid dynamics, we achieve on-the-fly policy optimization, effectively turning limited flight data into sustained performance gains.}

Through extensive real-world experiments, we demonstrate that our framework allows a quadrotor to reliably learn agile maneuvers near the saturation limit of the CTBR command interface. Empirically, the system evolves a conservative base policy from a peak speed of \SI{2.0}{\meter\per\second} to \SI{7.3}{\meter\per\second} within approximately \SI{100}{\second} of flight time. Beyond pure agility, we demonstrate the practical utility of our method in a multi-point inspection mission under strong external wind disturbances (Fig.~\ref{fig:catcheye}). Notably, after only \SI{2}{\minute} of real-world training, the quadrotor reduces mission completion time by {42\%} while maintaining low tracking error under unknown aerodynamic disturbances. These results underscore that real-world adaptation serves not merely to compensate for modeling errors, but as a practical mechanism for aggressive performance improvement in dynamic environments.

%% file: sections/related_work.tex
\section{Related Work}
\subsection{Learning-based Agile Flight Control}
In recent years, learning-based approaches have gained significant attention for agile quadrotor flight control, owing to their ability to handle highly nonlinear dynamics and operate under aggressive flight conditions. 
Early work demonstrated the feasibility of applying reinforcement learning to quadrotor control, enabling waypoint tracking and recovery from challenging initial conditions~\cite{hwangbo2017control}.
Subsequent research showed that reinforcement learning can support highly agile behaviors, with RL-based controllers achieving state-of-the-art performance in autonomous drone racing~\cite{kaufmann23champion, Song23Reaching}. 
In these settings, learning-based controllers have even surpassed expert human performance in competitive racing scenarios~\cite{kaufmann23champion, Song23Reaching}.
Advances in training efficiency further indicate that, under carefully optimized pipelines, reinforcement learning policies for quadrotor control can be trained within seconds~\cite{eschmann2024learning}.
Beyond reinforcement learning, imitation learning has played an important role in accelerating training and improving stability for agile flight.
By leveraging expert demonstrations, imitation learning has been applied both as a standalone approach and as an initialization or regularization strategy for reinforcement learning, substantially improving sample efficiency and overall control performance~\cite{xing2024bootstrapping, xing2024contrastive}. 
Despite these impressive results, learning-based quadrotor controllers are typically trained offline and remain fixed during deployment, limiting their ability to adapt to changing real-world conditions.
\input{figures/pipeline}
\subsection{Efficient Control Policy Learning}
Traditional reinforcement learning methods often require extensive training time, though significant acceleration is possible through highly optimized physics simulation~\cite{eschmann2024learning}. 
Despite these improvements, these methods remain sample-inefficient due to the high variance associated with zeroth-order policy gradient estimation.
An alternative paradigm is policy learning via differentiable simulation, which exploits smooth and differentiable dynamics and reward formulations to compute \emph{first-order} policy gradients.
This enables substantially faster and more sample-efficient learning compared to standard RL methods~\cite{heeg2025learning}.
Differentiable simulation has been successfully applied to direct policy parameterizations, including parametric trajectory representations for underwater~\cite{nava2022fast} and sinusoidal control policies for robotic cutting tasks~\cite{heiden2021disect}. 
However, extending these techniques to neural network policies remains challenging. In practice, unstable gradients often limit applicability to short-horizon tasks with simplified contact dynamics and restricted variation in initial conditions~\cite{xu2023efficient}.
To address these limitations, prior work has explored a range of stabilizing techniques, including early termination of simulations at contact events, truncated backpropagation through time~\cite{xu2022accelerated}, and reward shaping or augmentation with learned critics~\cite{georgiev2024adaptive, luo2024residual}.
{However, these methods have yet to effectively address the challenge of safely pushing the platform toward its agility envelope to enable online learning of agile flight.}

%% file: figures/pipeline.tex
\begin{figure*}
    \centering
    \includegraphics[width=0.95\textwidth]{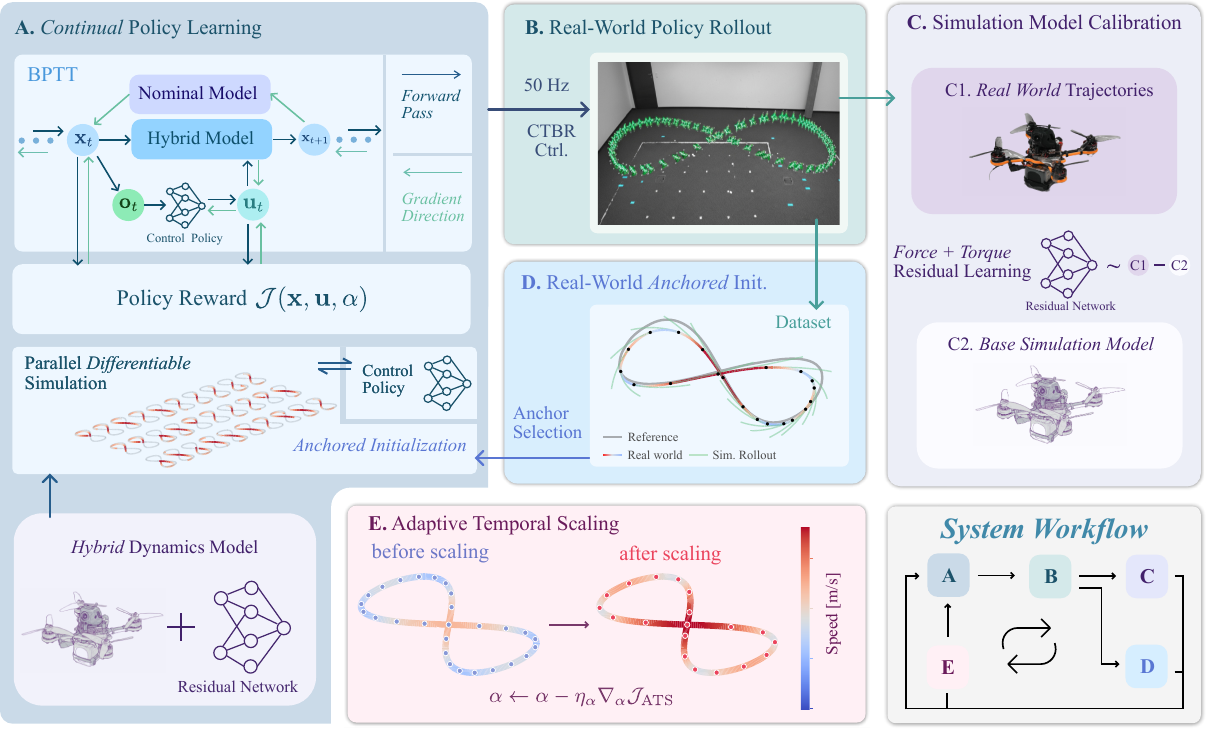}
\caption{\textbf{Overview of the self-adaptive autonomous flight framework.} The system operates as a continuous closed-loop cycle (bottom right) bridging physical execution and differentiable simulation:
  \textbf{(A) Policy Learning:} Leveraging a learned hybrid dynamics model in a differentiable simulator, we perform \textbf{RASH-BPTT} to optimize the control policy via massively parallelized rollouts.
  \textbf{(B) Real-World Rollout:} The agent executes the current policy on the physical quadrotor to collect state-action-transition data.
  \textbf{(C) Model Calibration:} Collected data is used to update the \textit{Hybrid Dynamics Model} online, where a neural residual network learns to compensate for the reality gap (e.g., unmodeled aerodynamics, delays) of the nominal rigid-body model.
  \textbf{(D) Anchored Initialization:} To mitigate compounding prediction errors, simulation rollouts are initialized (\textit{anchored}) using the most recent real-world state estimates rather than random resets.
\textbf{(E) Adaptive Temporal Scaling (ATS):} Integrated directly with policy optimization, the trajectory time-scale $\alpha$ is jointly optimized using real-world rollouts. By leveraging analytical gradients derived from closed-loop sensitivity, ATS maximizes agility while enforcing safety constraints via a barrier function.}
    \label{fig:pipeline}
    \vspace{-0.6cm}
\end{figure*}

%% file: sections/methodology.tex
\section{Methodology}

In this section, we introduce our self-adaptive framework. As illustrated in Fig.~\ref{fig:pipeline}, the core workflow operates as a continuous closed-loop cycle bridging physical execution and differentiable simulation.
The process begins with an initialization phase, where we employ standard BPTT to train a base policy on the nominal quadrotor model. Benefiting from the sample efficiency of differentiable simulation, this pre-training typically converges within seconds. Once deployed, the agent commences the real-world adaptation loop. First, the system collects flight data to update the \textit{Hybrid Dynamics Model}, using a neural residual network to capture both internal dynamic discrepancies and external disturbances. Subsequently, leveraging this calibrated model, we execute \textit{RASH-BPTT} to optimize the policy via massively parallelized rollouts. To mitigate compounding prediction errors, these rollouts utilize \textit{Anchored Initialization}, resetting the simulation to the most recent real-world state estimates. Tightly coupled with policy optimization, \textit{Adaptive Temporal Scaling (ATS)} is applied to dynamically adjust the reference trajectory evolution. By optimizing a learnable parameter $\alpha$ through analytical gradients derived from closed-loop sensitivity directly on real-world rollouts, the system maximizes agility while maintaining safety constraints.

The framework consists of three hierarchical components:
\begin{enumerate}
    \item \textit{Online Residual Learning}: Calibrates the online hybrid dynamics model to compensate for the reality gap using real-world data (Section~\ref{subsec:residual_learning}).
    \item \textit{RASH-BPTT}: Performs online policy optimization via differentiable simulation, anchored to the current physical state to ensure robust updates (Section~\ref{subsec:rash_bptt}).
    \item \textit{Adaptive Temporal Scaling (ATS)}: Jointly optimizes the reference trajectory's time evolution to balance flight agility and safety (Section~\ref{subsec:ats}).
\end{enumerate}

\subsection{Online Residual Dynamics Learning}
\label{subsec:residual_learning}

To avoid precise offline system identification and to adapt to external disturbances and platform-specific discrepancies, we employ a hybrid dynamics model.
Specifically, we learn a neural residual term that augments the nominal rigid-body dynamics and is trained by minimizing one-step state prediction error, inspired by Neural ODEs~\cite{chen2018neural}.

\subsubsection{Hybrid Continuous-Time Dynamics}

We define the quadrotor state as
\[
\mathbf{x} \triangleq (\mathbf{p}, \mathbf{v}, \mathbf{R})
\in \mathbb{R}^3 \times \mathbb{R}^3 \times \mathrm{SO}(3),
\]
where $\mathbf{p}$ and $\mathbf{v}$ are the position and velocity expressed in the world frame, and
$\mathbf{R}$ maps vectors from the body frame to the world frame.
Let $\mathbf{e}_3 \triangleq [0,0,1]^\top$ denote the third canonical basis vector; thus,
$\mathbf{R}\mathbf{e}_3$ is the body-frame thrust axis expressed in the world frame.
Throughout the paper, gravity is taken to act along $-g\mathbf{e}_3$.
The control input is a collective-thrust/body-rate (CTBR) command
$
\mathbf{u} \triangleq
\begin{bmatrix}
c & \bm{\omega}_{\mathrm{cmd}}^\top
\end{bmatrix}^{\top}
\in \mathcal{U},
$
with admissible input set
\[
\mathcal{U}
\triangleq
[c_{\min},c_{\max}]
\times
\left\{
\bm{\omega}\in\mathbb{R}^3
\,\middle|\,
\|\bm{\omega}\|_{\infty}\leq \omega_{\max}
\right\}.
\]
Here, $c$ is the mass-normalized collective thrust, with units of $\mathrm{m\,s^{-2}}$, and
$\bm{\omega}_{\mathrm{cmd}}\in\mathbb{R}^3$ is the commanded body rate.
The infinity-norm constraint imposes an axis-wise saturation on the commanded body rate.
We use the CTBR as an abstraction over individual rotor thrusts: it simplifies policy learning and leverages a well-tuned low-level rate controller, while making the bound on
$\|\bm{\omega}_{\mathrm{cmd}}\|_{\infty}$ a controller-level command ceiling rather than a fundamental hardware limit.


We model the continuous-time quadrotor dynamics as a nominal CTBR model
augmented with neural residual terms:
\begin{subequations}
\label{eq:continuous_ode}
\begin{align}
    \dot{\mathbf{p}} &= \mathbf{v}, \\
    \dot{\mathbf{v}} &=
    c\,\mathbf{R}\mathbf{e}_3
    - g\mathbf{e}_3
    + \mathbf{a}_{\mathrm{res}}(\bm{\zeta};\theta), \\
    \dot{\mathbf{R}} &=
    \mathbf{R}
    \left(
    \bm{\omega}_{\mathrm{cmd}}
    + \bm{\omega}_{\mathrm{res}}(\bm{\zeta};\theta)
    \right)^\wedge .
\end{align}
\end{subequations}
This defines the hybrid vector field
$\mathbf{f}_{\mathrm{hybrid}}(\mathbf{x},\mathbf{u};\theta)$.
Here, $(\cdot)^\wedge:\mathbb{R}^3\to\mathfrak{so}(3)$ denotes the
skew-symmetric map to the Lie algebra $\mathfrak{so}(3)$,
$\bm{\zeta}$ denotes the residual-model input features, and $\theta$ denotes
the residual-model parameters.
The residual terms
$\mathbf{a}_{\mathrm{res}}\in\mathbb{R}^3$ and
$\bm{\omega}_{\mathrm{res}}\in\mathbb{R}^3$
capture unmodeled translational accelerations and body-rate discrepancies,
respectively.
This formulation assumes that the low-level body-rate loop tracks CTBR commands
sufficiently fast, so that its dominant tracking errors can be represented as
additive residuals on the commanded body rates.

The residual network $\mathbf f_{\mathrm{res}}(\bm\zeta;\theta)$ takes the feature vector
\begin{equation}
\bm\zeta \triangleq [\mathbf p^\top,\ \mathbf v^\top,\ \mathrm{vec}(\mathbf R)^\top,\ \mathbf u^\top]^\top,
\end{equation}
where $\mathrm{vec}(\cdot)$ denotes column-stacking vectorization, and outputs
$\mathbf f_{\mathrm{res}}(\bm\zeta;\theta) = \big[\mathbf a_{\mathrm{res}}^\top,\ \bm\omega_{\mathrm{res}}^\top\big]^\top$.

\subsubsection{Differentiable Integration and Online Training}

To obtain a discrete-time transition, we employ a differentiable fourth-order Runge--Kutta (RK4) integrator $\Phi_{\mathrm{RK4}}$.
Beyond improved numerical accuracy, RK4 is important for our one-step residual learning: its intermediate stages partially advance $\mathbf R$ within the step, making the translational prediction $(\mathbf p_{k+1},\mathbf v_{k+1})$ sensitive to $\bm\omega_{\mathrm{res}}$ and thus allowing position/velocity errors to supervise attitude-related residuals.
We define the discrete dynamics map
\begin{equation}
    \mathbf x_{k+1}
    =
    \mathbf F_{\Delta t}(\mathbf x_k,\mathbf u_k;\theta)
    \triangleq
    \Phi_{\mathrm{RK4}}\!\left(\mathbf x_k,\mathbf u_k,\mathbf f_{\mathrm{hybrid}};\Delta t\right).
    \label{eq:discrete_dynamics}
\end{equation}
In implementation, $\Phi_{\mathrm{RK4}}$ is realized as a Lie-group integrator that updates $\mathbf R$ on $SO(3)$ using the exponential map, which prevents attitude drift and preserves $\mathbf R\in SO(3)$ while remaining fully differentiable.

We optimize parameters $\theta$ online using a sliding-window replay buffer $\mathcal{B}$ of recent transitions $(\mathbf{x}_k, \mathbf{u}_k, \mathbf{x}_{k+1})$ collected from the state estimator.
Instead of explicitly supervising acceleration targets (which are often noisy), we minimize the \emph{integrated} one-step prediction error between the simulated next state $\hat{\mathbf{x}}_{k+1} = \mathbf F_{\Delta t}(\mathbf x_k,\mathbf u_k;\theta)$ and the measured state $\mathbf{x}_{k+1}$:
\begin{equation}
    \mathcal{L}_{\text{res}}(\theta) = \frac{1}{|\mathcal{B}|} \sum_{k \in \mathcal{B}} \mathcal{D}({\mathbf{x}}_{k+1}, \hat{\mathbf{x}}_{k+1}) 
    + \lambda_{\text{reg}} \sum_{l} \|\mathbf{W}_l\|_\sigma.
    \label{eq:residual_loss}
\end{equation}
{The discrepancy metric $\mathcal{D}(\cdot, \cdot)$ combines the Euclidean translation error and the geodesic distance on the rotation manifold. The rotational term is formulated as $\|\mathrm{Log}(\hat{\mathbf{R}}^\top \mathbf{R})^\vee\|_2^2$, where $\mathrm{Log}: SO(3) \rightarrow \mathfrak{so}(3)$ denotes the logarithmic map to the Lie algebra, and $(\cdot)^\vee$ maps the skew-symmetric matrix to its corresponding vector in $\mathbb{R}^3$. This ensures the loss strictly respects the geometry of $SO(3)$.}
The regularization term penalizes the spectral norm $\|\mathbf{W}_l\|_\sigma$ of each network layer. As motivated by~\cite{shi2019neural}, constraining the Lipschitz constant of the residual dynamics is crucial for bounding gradient variance and ensuring numerical stability during long-horizon recursive optimization.

\subsection{Real-World Anchored Short-Horizon BPTT}
\label{subsec:rash_bptt}

Building on the learned hybrid dynamics, we propose \textit{Real-World Anchored Short-Horizon BPTT (RASH-BPTT)}. To address \emph{model exploitation} caused by compounding errors~\cite{janner2019trust}, our method \emph{anchors} each rollout to the instantaneous real-world state. Furthermore, by restricting optimization to a \emph{short horizon} (Fig.~\ref{fig:pipeline}D), we prevent error accumulation, ensuring that policy gradients remain grounded in reality and optimized specifically for the agent's current dynamic conditions.

We parameterize the control policy $\pi_\phi$ as a multilayer perceptron (MLP) with two hidden layers of 256 units. The policy takes an observation vector $\mathbf o_k = [{\mathbf x}_k, \mathbf x_{\mathrm{ref},k}, \mathbf h_k]$, composed of the current state, the reference trajectory state, and a history of past actions $\mathbf h_k$ to encourage control smoothness. The output is a four-dimensional CTBR command $\mathbf u_k \in \mathbb{R}^4$.

At each real-world time step $t$, we initialize the differentiable rollout with the current state estimate ${\mathbf x}_0 := \mathbf x_{\mathrm{real}}(t)$. The policy generates control actions conditioned on the current rollout state $\mathbf u_k := \pi_\phi(\mathbf o_k)$. We then unroll the short-horizon dynamics for $H$ steps using the learned discrete map from Eq.~\eqref{eq:discrete_dynamics}:
\begin{equation}
    {\mathbf x}_{k+1} = \mathbf F_{\Delta t}({\mathbf x}_k,\mathbf u_k;\theta), \quad k=0,\dots,H-1.
    \label{eq:rash_rollout}
\end{equation}

The optimization objective is to maximize the cumulative discounted reward along this predicted trajectory:
\begin{equation}
    \mathcal J(\phi) = \sum_{k=0}^{H-1} \gamma^k\, r({\mathbf{x}}_k, \mathbf{u}_k, \mathbf x_{\mathrm{ref},k}),
    \label{eq:policy_objective}
\end{equation}
where $\gamma\in(0,1]$ is the discount factor and $r(\cdot)$ is the task reward function. 
We obtain the gradient $\nabla_\phi \mathcal J$ by performing BPTT through the unrolled computation graph. This is implemented via JAX's automatic differentiation engine~\cite{jax2018github}, allowing us to differentiate through the RK4 integrator and neural residuals exactly.
The policy parameters are updated via stochastic gradient ascent: $\phi \leftarrow \phi + \eta_{\pi} \nabla_\phi \mathcal J$. 
By combining autodiff with JAX's Just-In-Time (JIT) compilation and automatic vectorization (vmap), we achieve highly efficient, parallelized updates on the GPU.

\subsection{Adaptive Temporal Scaling via Differentiable Optimization}
\label{subsec:ats}

While RASH-BPTT optimizes the control policy parameters, achieving time-optimal performance under varying conditions requires dynamically adapting the reference trajectory's evolution. To this end, we introduce \textit{Adaptive Temporal Scaling (ATS)}, a closed-loop mechanism that decouples spatial path planning from the temporal execution profile. 
This approach gives the system a form of ``temporal elasticity'': it aggressively compresses the time domain (reducing $\alpha$) to maximize agility when tracking is precise, yet automatically relaxes kinematic demands (increasing $\alpha$) to restore stability upon encountering significant disturbances or model mismatches. 
Unlike heuristic speed adjustment rules, ATS is formulated as a holistic online optimization problem. Inspired by~\cite{nan2025efficient}, we explicitly minimize the future discrepancy between the \textit{real} and \textit{reference} trajectories, utilizing the differentiable hybrid model as a proxy to anticipate the closed-loop effects of time-scale adjustments.

\subsubsection{Parameterization}
We parameterize the reference trajectory as a piecewise polynomial sequence. For each segment, the spatial path is defined by a coefficient matrix $\mathbf{C} \in \mathbb{R}^{3 \times N}$ (representing a polynomial of degree $N-1$ in 3D space) and a temporal basis vector $\bm{\beta}(t) = [1, t, t^2, \cdots, t^{N-1}]^\top$. 
To control the execution rate, we introduce a scalar scaling parameter $\alpha \in \mathbb{R}_{>0}$. 
The reference position $\mathbf{p}_{\text{ref}}$ at a specific trajectory time $\tau$ is formulated as:
\begin{equation}
    \mathbf{p}_{\text{ref}}(\tau; \alpha) = \mathbf{C} \bm{\beta}\left(\frac{\tau}{\alpha}\right).
    \label{eq:poly_traj}
\end{equation}
Here, $\alpha$ serves as a \textit{time dilation factor}. By exploiting the differential flatness property~\cite{faessler2017differential}, we map the derivatives to the full reference state $\mathbf{x}_{\text{ref}}(\tau; \alpha)$.

\subsubsection{Optimization Objective and Update}

Direct optimization of $\alpha$ using real-world rollouts is intractable, as the physical system execution is non-differentiable and does not provide analytical gradients.
To address this, we employ the differentiable hybrid model as a \textit{proxy} to perform \textit{counterfactual inference}~\cite{buesing2018woulda}.
At each update step, given a recent real-world rollout $\{(\bar{\mathbf x}_k,\bar{\mathbf u}_k)\}_{k=0}^{H-1}$, we analyze the sensitivity of the closed-loop performance to the time-scale $\alpha$ by constructing a \emph{counterfactual} state sequence $\hat{\mathbf x}_k(\alpha)$.
This allows us to formulate a composite potential function $\mathcal J_{\text{ATS}}$ over the horizon $H$ that balances execution speed with safety:
\begin{equation}
    \mathcal J_{\text{ATS}}(\alpha)
    =
    \lambda_{\mathrm{speed}}\, \alpha
    +\lambda_{\mathrm{safe}}\sum_{k=0}^{H-1}\Psi\!\left(\mathcal{E}_k(\alpha)-\mathcal{E}_{\mathrm{th}}\right),
    \label{eq:potential}
\end{equation}
where the error metric is defined as 
\begin{equation}
    \mathcal{E}_k(\alpha) \triangleq \mathcal{D}(\hat{\mathbf x}_k(\alpha),\, \mathbf{x}_{\mathrm{ref},k}(\alpha)).
\end{equation}
Here, $\mathcal{D}(\cdot,\cdot)$ denotes the geometry-aware discrepancy from Eq.~\eqref{eq:residual_loss}, and $\mathcal E_{\text{th}}$ is a predefined safety threshold.

Crucially, the counterfactual state $\hat{\mathbf x}_k(\alpha)$ serves as a differentiable surrogate for the physical state.
It is anchored to the real observations, satisfying $\hat{\mathbf x}_k(\alpha_0)=\bar{\mathbf x}_k$ at the current time-scale $\alpha_0$.
However, unlike the fixed real history $\bar{\mathbf x}_k$, the evolution of $\hat{\mathbf x}_k(\alpha)$ depends on $\alpha$. Its local sensitivity is approximated by linearizing the hybrid model dynamics along the real rollout $\{(\bar{\mathbf x}_k,\bar{\mathbf u}_k)\}$.
Here, $\lambda_{\mathrm{speed}}$ and $\lambda_{\mathrm{safe}}$ are weighting factors, and $\Psi(z) = \frac{1}{\kappa}\ln(1 + \exp(\kappa z))$ is the Softplus barrier with sharpness $\kappa$.

\input{figures/barrier}

We then compute the gradient $\nabla_\alpha \mathcal J_{\text{ATS}}$ using {closed-loop sensitivity} analysis. By applying the chain rule, the gradient combines the analytic gradient of the reference with the recursive sensitivity of the system state, approximated by the hybrid model dynamics $\mathbf{F}_{\Delta t}$:
\begin{equation}
\frac{{\rm d} \mathcal J_{\text{ATS}}}{{\rm d} \alpha}
=
\lambda_{\mathrm{speed}}
+\lambda_{\mathrm{safe}}\sum_{k=0}^{H-1}
\Psi'\!\left(\mathcal{E}_k(\alpha)-\mathcal{E}_{\mathrm{th}}\right)\,
\frac{{\rm d}\mathcal{E}_k(\alpha)}{{\rm d}\alpha},
\end{equation}
where
\begin{equation}
\frac{{\rm d}\mathcal{E}_k}{{\rm d}\alpha}
=
\frac{\partial \mathcal{E}_k}{\partial \mathbf{x}_{\text{ref},k}}
\frac{\partial \mathbf{x}_{\text{ref},k}}{\partial \alpha}
+
\frac{\partial \mathcal{E}_k}{\partial \hat{\mathbf{x}}_k}\mathbf{S}_k.
\end{equation}
Here, $\mathbf{S}_k \triangleq \frac{{\rm d} \hat{\mathbf x}_k}{{\rm d} \alpha}$ is the closed-loop state sensitivity, with $\mathbf{S}_0=\mathbf{0}$, and evolves as:
\begin{equation}
    \mathbf{S}_{k+1}
    =
    \mathbf{A}_k\,\mathbf{S}_k
    +\mathbf{B}_k\,\frac{{\rm d} \mathbf{u}_k}{{\rm d} \alpha},
    \label{eq:state_sens}
\end{equation}
with Jacobians evaluated on the real rollout:
\begin{equation}
\mathbf{A}_k \triangleq \left.\frac{\partial \mathbf{F}_{\Delta t}}{\partial \mathbf{x}}\right|_{(\bar{\mathbf x}_k,\bar{\mathbf u}_k)},
\mathbf{B}_k \triangleq \left.\frac{\partial \mathbf{F}_{\Delta t}}{\partial \mathbf{u}}\right|_{(\bar{\mathbf x}_k,\bar{\mathbf u}_k)}.
\end{equation}
The action sensitivity is obtained by differentiating through the policy:
\begin{equation}
    \frac{{\rm d}\mathbf{u}_k}{{\rm d}\alpha}
    =
    \frac{\partial \pi_\phi}{\partial \mathbf{o}_k}
    \left(
        \frac{\partial \mathbf{o}_k}{\partial \hat{\mathbf x}_k}\mathbf{S}_k
        +
        \frac{\partial \mathbf{o}_k}{\partial \mathbf{x}_{\mathrm{ref},k}}
        \frac{\partial \mathbf{x}_{\mathrm{ref},k}}{\partial \alpha}
    \right),
    \label{eq:action_sensitivity}
\end{equation}
where the observation $\mathbf{o}_k$ is a function of the state $\hat{\mathbf x}_k$ {and} the reference $\mathbf x_{\mathrm{ref},k}(\alpha)$. While the functional dependency on $\hat{\mathbf x}_k$ allows for differentiation, all Jacobian matrices (e.g., $\frac{\partial \pi_\phi}{\partial \mathbf{o}_k}, \frac{\partial \mathbf{o}_k}{\partial \hat{\mathbf x}_k}$) are evaluated at the nominal operating point defined by the real rollout $(\bar{\mathbf x}_k, \bar{\mathbf u}_k)$.
Additional dependencies in $\mathbf{o}_k$ (e.g., action history) can be included analogously and are handled automatically in our autodiff implementation.

The parameter is updated via projected gradient descent:
\begin{equation}
    \alpha
    \leftarrow
    \Pi_{[\alpha_{\min},\,\alpha_{\max}]}\!(\alpha - \eta_{\alpha}\nabla_\alpha\mathcal J_{\text{ATS}}).
\end{equation}
This update rule naturally maintains an equilibrium near the safety boundary, as visualized in Fig.~\ref{fig:barrier_func}. In the \textit{safe regime} ($\mathcal{E} \ll \mathcal{E}_{\mathrm{th}}$), the barrier gradient vanishes, and the linear term $\lambda_{\text{speed}}$ drives $\alpha$ down to accelerate execution. Conversely, in the \textit{unsafe regime}, the barrier gradient dominates, using the model-based sensitivity to increase $\alpha$ and restore feasibility.

%% file: figures/barrier.tex
\begin{figure}[t]
    \centering
    \includegraphics[width=0.43\textwidth]{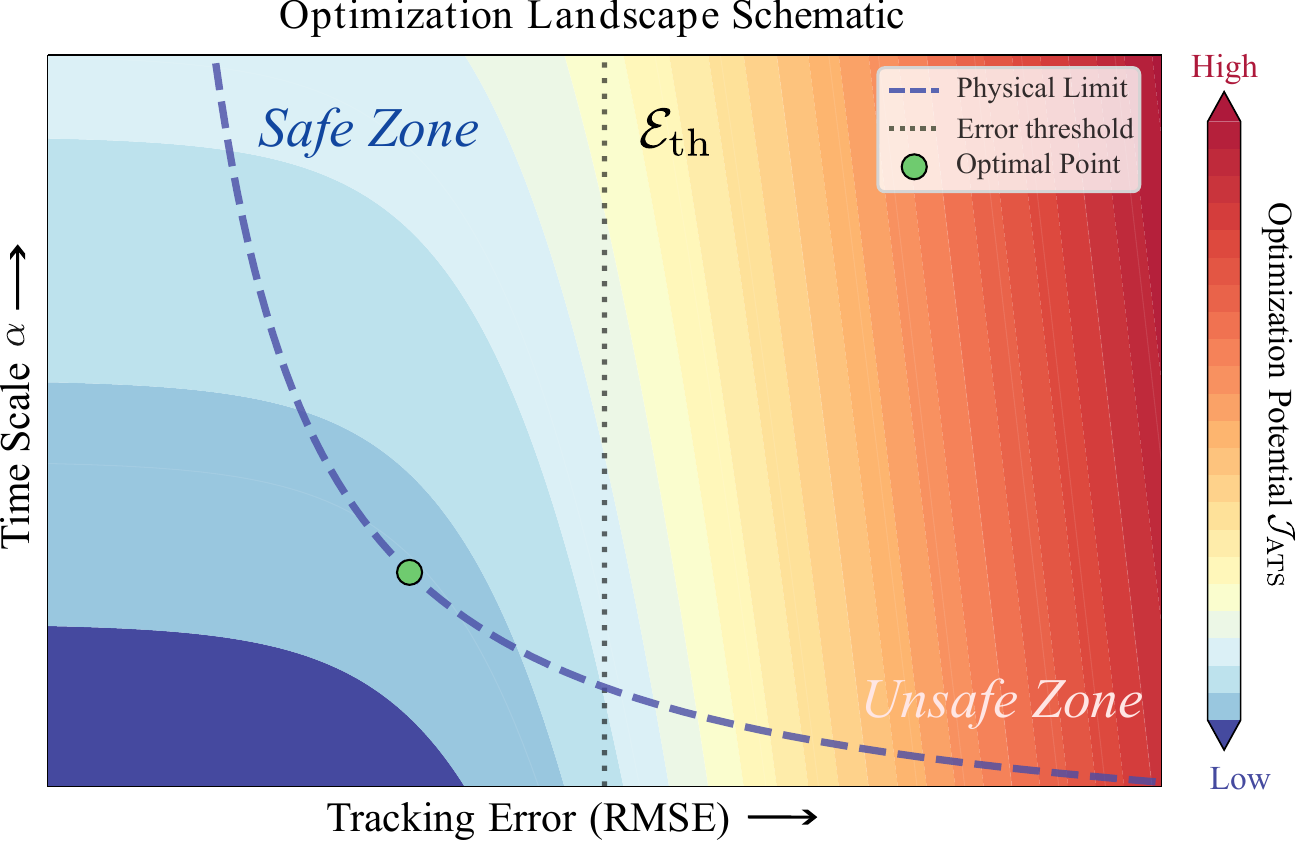}
    \caption{\textbf{The optimization landscape for ATS.}
    The heatmap visualizes the composite potential $\mathcal{J}_{\text{ATS}}$, balancing agility (low $\alpha$) against safety. The landscape transitions from the Safe Zone (blue) to the Unsafe Zone (red) determined by the tracking error threshold $\mathcal{E}_{\text{th}}$ (dotted line) and a {schematic representation} of the system's physical limits (dashed curve). The green dot marks the optimal equilibrium: the most aggressive time scale achievable within safe tracking bounds.}
    \label{fig:barrier_func}
    \vspace{-0.7cm} 
\end{figure}

%% file: sections/experiments.tex
\input{figures/disturbance}

\input{figures/bodyrates}

\section{Results and Experiments}\label{sec:results_and_experiments}

In this section, we present comprehensive experiments to validate the effectiveness of our self-adaptive framework. We conduct both real-world flight tests and extensive simulation comparisons to answer the following research questions:
\begin{itemize}
    \item \textit{RQ1 (Agility):} Can the proposed framework push the quadrotor to the saturation limit of its CTBR command interface in real time, starting from a conservative policy?
    \item \textit{RQ2 (Robustness):} Is the framework capable of adapting to internal dynamic changes (e.g., payload, damage) and unknown external disturbances (e.g., wind) while maintaining safety?
    \item \textit{RQ3 (Ablation):} {How do the individual components, residual learning and real-world anchored initialization, contribute to the overall performance?}
\end{itemize}

We validate the proposed framework entirely in real-world flight using a quadrotor based on the Agilicious platform~\cite{foehn2022agilicious}, as shown in Fig.~\ref{fig:disturbance}.
A motion-capture system provides state estimates at \SI{100}{\hertz}.
An off-board workstation runs the online learning loop and sends commands to the onboard flight controller at \SI{50}{\hertz}.
{The workstation is equipped with an NVIDIA RTX 4090 GPU; using JAX with JIT compilation and \texttt{vmap}, each adaptation step (100 epochs: $\sim$3\,s residual + $\sim$6\,s policy) completes in $\sim$9\,s. The 50\,Hz command loop requires only a forward pass through a lightweight MLP ($\sim$100\,$\mu$s), so policy execution is decoupled from learning.}

\subsection{Real-World Agile Flight and Adaptation}

\subsubsection{Approaching the CTBR Interface Limit (RQ1)}

We evaluate whether our framework can adapt from conservative flight to
near-limit agility on two canonical trajectories: a linear shuttle and a
Figure-8. As shown in Fig.~\ref{fig:catcheye}, on the \emph{Linear Shuttle},
the vehicle starts from a conservative maximum speed of
$2.0~\mathrm{m/s}$ and reaches $7.3~\mathrm{m/s}$ after roughly
$100~\mathrm{s}$ of in-flight adaptation. On the \emph{Figure-8}, the speed
similarly increases from $2.2~\mathrm{m/s}$ to $7.2~\mathrm{m/s}$ over the
same adaptation horizon.

Across both trajectories, the adapted policies reach the axis-wise saturation
limit of the CTBR body-rate command interface. Fig.~\ref{fig:bodyrates} shows
the commanded body rate $\bm{\omega}_{\mathrm{cmd}}$, where the raw commands
and their 2\,s sliding-window mean indicate persistent saturation at the input
bound, $\|\bm{\omega}_{\mathrm{cmd}}\|_{\infty} = 6~\mathrm{rad/s}$,
during the later stages of adaptation. This indicates that ATS does not merely
improve tracking at a fixed aggressiveness level; rather, it safely exploits the
improved predictive accuracy provided by residual learning and RASH-BPTT to
push the trajectory toward the boundary of the CTBR command regime.

To contextualize the achieved lap times, we compute two zero-delay offline
lower bounds via trajectory optimization: a \emph{CTBR bound} under ideal
rigid-body dynamics with the same command limits
$c \leq 25~\mathrm{m\,s^{-2}}$ and
$\|\bm{\omega}_{\mathrm{cmd}}\|_{\infty}\leq 6~\mathrm{rad/s}$, and a
\emph{rotor-thrust bound} that directly constrains individual motor thrusts
without the CTBR body-rate ceiling. Table~\ref{tab:theoretical_bounds}
summarizes the comparison.

\begin{table}[t]
    \centering
    \caption{Real-world lap times versus zero-delay offline lower bounds.}
    \label{tab:theoretical_bounds}
    \small
    \setlength{\tabcolsep}{4pt}
    \begin{tabular}{@{}lcc@{}}
        \toprule
        Method & Figure-8 (s) & Shuttle (s) \\
        \midrule
        Rotor-thrust bound$^{\dagger}$ & 2.02 & 1.42 \\
        CTBR bound                      & 2.24 & 1.49 \\
        Ours                            & 2.60 & 2.30 \\
        \bottomrule
    \end{tabular}

    \vspace{2pt}
    \begin{minipage}{0.95\columnwidth}
        \footnotesize
        \raggedright
        $^{\dagger}$Not reachable through the CTBR interface; the optimized
        trajectory requires body rates exceeding $28~\mathrm{rad/s}$.
    \end{minipage}
\end{table}

Our Figure-8 result, $2.60~\mathrm{s}$, approaches the CTBR bound of
$2.24~\mathrm{s}$, indicating operation near the command-interface ceiling.
The larger shuttle gap, $2.30~\mathrm{s}$ versus $1.49~\mathrm{s}$, reflects
the trajectory's stronger sensitivity to actuation delays, rapid acceleration
reversals, and aerodynamic effects absent from the zero-delay reference.

\input{figures/broken_add_weight}

\subsubsection{Robustness to Hardware Variations (RQ2)}
To evaluate the system's resilience to internal dynamic shifts, we conducted flight tests on the Figure-8 trajectory under three distinct hardware degradations: \textit{Added Mass} (adding a \SI{60}{\gram} payload), \textit{Propeller Damage} (clipping propeller tips), and \textit{Combined} (both mass increase and propeller damage).

As illustrated in Fig.~\ref{fig:broken_add_weight}, compared to the intact baseline, the achievable maximum speeds in all three scenarios naturally decreased, reflecting the reduced physical authority of the compromised hardware. {In contrast,} our framework successfully identified the \textit{new} feasible operating envelope for each configuration, allowing the drone to operate at the boundary of CTBR body-rate saturation while maintaining stability.

A key observation underscores the efficacy of our Residual Dynamics Learning. Immediately following hardware alteration, the nominal policy exhibited significant tracking errors due to severe model mismatch. Remarkably, after just a \textit{single update iteration} of the residual model, the tracking error dropped precipitously. This rapid adaptation enabled ATS to confidently decrease $\alpha$, thereby compressing the time scale and increasing agility. This confirms that the residual network effectively captures platform-specific {properties}, enabling the ATS+RASH-BPTT loop to safely explore the unknown performance envelope.

\subsubsection{Application: Time-Optimal Inspection under Unknown Disturbances}
\label{subsubsec:application}

To demonstrate practical utility, we evaluated the framework in a realistic multi-stage inspection mission subject to unknown wind fields between landmarks (Fig.~\ref{fig:catcheye}).
Despite turbulent aerodynamic disturbances, the system successfully maintained stable tracking via online residual adaptation. This robustness empowered the ATS mechanism to progressively compress the execution duration, ultimately reducing the total motion time by {42\%} (from \SI{12}{\second} to \SI{7}{\second}) while adhering to safety constraints, as visualized in Fig.~\ref{fig:inspection}.

\input{figures/inspection}

\input{figures/ablation}
\input{figures/fig8_wind}

\subsection{Simulation Evaluation}
\subsubsection{Ablation Study}
{While our real-world experiments demonstrated the system's overall efficacy, we use simulation to disentangle the contributions of individual modules.
    Since no existing method supports online learning while pushing toward the agility envelope in real-world agile-flight settings, we integrate our ATS framework into all baselines to ensure a fair comparison.
    Consequently, the comparisons below represent ablated variants of our holistic system, evaluated on tracking accuracy (RMSE) and the maximum speed optimized by ATS:
}

\begin{itemize}
    \item \textit{Baseline:} Relies solely on the \emph{nominal} dynamics model optimized via standard BPTT, serving as a lower bound to benchmark the efficacy of residual learning.

    \item \textit{LOTF}~\cite{pan2026learning}: Adapts the LOTF architecture to our framework. It models residual \emph{linear acceleration} but omits rotational dynamics.

    \item \textit{Anchor Only:} A variant that applies RASH-BPTT with anchored rollouts while disabling the learned residual.

    \item \textit{Residual Only:} Incorporates the full hybrid residual model (linear acceleration and body rates) but reverts to standard BPTT optimization.

    \item \textit{Ours:} The complete approach that integrates the full hybrid residual dynamics with anchored rollouts for optimal performance.
\end{itemize}

We set the tracking error threshold $\mathcal{E}_{\text{th}}$ to $0.35$\,m. The results, summarized in Fig.~\ref{fig:ablations}, yield three key insights:

\textit{{Residual learning lifts the agility ceiling that ATS alone cannot break through.}}
Across all methods, incorporating residual dynamics substantially reduces tracking errors. Since the ATS mechanism allows for time compression (decreasing $\alpha$) \textit{only} when the predicted error remains within the safety margin, lower RMSE directly translates into more aggressive time-scaling capabilities. Without residuals, the baseline hits the error threshold early, effectively capping the maximum speed.

\textit{Angular dynamics fidelity is critical in high-speed regimes.}
Compared to LOTF (acceleration residuals only), our full residual model (acceleration + angular velocity) achieves superior tracking accuracy as speed increases. This indicates that compensating for rotational dynamics mismatches (e.g., aerodynamic drag moments) is essential for maintaining prediction fidelity during aggressive maneuvers, which in turn empowers ATS to push performance boundaries further.

\textit{Anchored rollouts enhance optimization stability.}
Comparing \textit{Anchor Only} against standard initialization reveals that anchoring the simulation rollout to the current real-world state mitigates the \textit{distribution shift} between optimization and deployment. Practically, this keeps the optimization trajectory aligned with the on-policy region, yielding more stable convergence than optimizing from a reset reference state.

Overall, our full method achieves the optimal speed--accuracy trade-off, demonstrating a synergistic effect between residual dynamics (ensuring prediction accuracy) and anchored optimization (ensuring stable adaptation).

\subsubsection{Online Adaptation under Wind Disturbance}
We further evaluate robustness by benchmarking \textit{Ours} against the \textit{Baseline} in a simulated \SI{5}{\meter\per\second} wind field (Fig.~\ref{fig:fig8_wind}).
Within just a single update, the residual-enhanced policy reduces tracking RMSE by approximately a factor of four compared to the baseline. This rapid error suppression is critical: it creates the necessary safety margin for ATS to initiate aggressive time-scaling early in the flight, validating that accurate dynamics are a prerequisite for agility.
Over ten iterations, our method increases peak speed by a factor of 2.6, from \SIrange{2.6}{6.8}{\meter\per\second}, while maintaining a bounded tracking error below $\mathcal E_{\text{th}} = 0.35$\,m. In contrast, the baseline fails to accelerate meaningfully due to persistent model mismatch. This confirms that residual learning effectively compensates for external disturbances, empowering the ATS optimizer to safely push the system toward the saturation limit of the CTBR command interface.

%% file: figures/disturbance.tex
\begin{figure}[t!]
    \centering
    \includegraphics[width=0.48\textwidth]{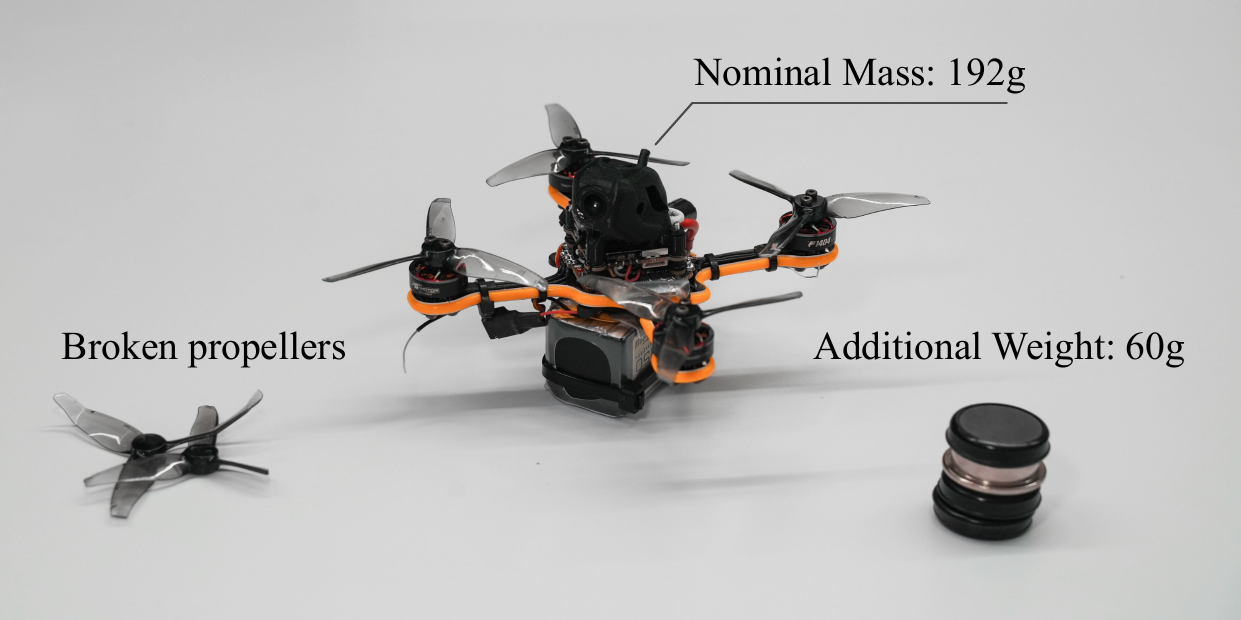}
    \caption{\textbf{Experimental platform with extreme modifications.} 
    To validate robustness, the nominal quadrotor (192\,g) is subjected to drastic degradations: {mechanically clipped propellers} (inducing aerodynamic loss) and a {60\,g payload}. This {31\% mass increase} significantly alters the inertial properties and reduces the thrust-to-weight ratio.}
    \label{fig:disturbance}
     \vspace{-0.6cm}
\end{figure}

%% file: figures/bodyrates.tex
\begin{figure}[t!]
    \centering
    \includegraphics[width=0.48\textwidth]{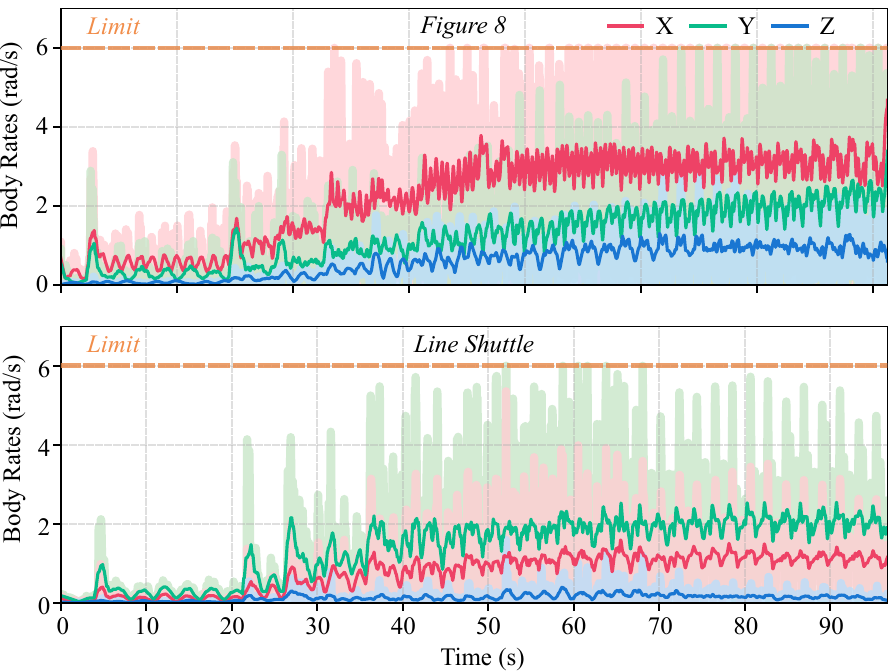}
    \caption{Commanded body-rate $\bm{\omega}_{\mathrm{cmd}}$ during real-world Figure-8 and Linear Shuttle flights. Solid curves denote the \SI{2}{\second} sliding-window mean, overlaid on raw measurements (background traces). The orange dashed line marks the actuation limit (\SI{6}{\radian\per\second}).}
    
    \label{fig:bodyrates}
    \vspace{-0.6cm}
\end{figure}

%% file: figures/broken_add_weight.tex
\begin{figure}[!t]
    \centering
    \includegraphics[width=0.48\textwidth]{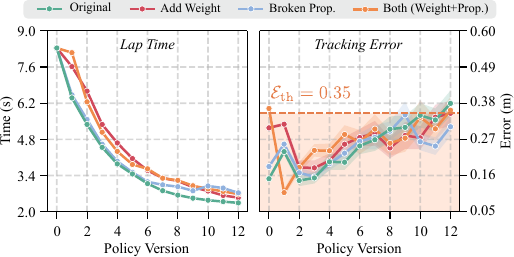}
    \caption{Adaptation to hardware variations. The framework is tested with (i) added mass, (ii) propeller damage, and (iii) combined conditions. In all cases, the residual learning module swiftly compensates for the dynamic mismatch within one iteration, enabling the ATS to safely push the compromised hardware to its new physical limits.}
    \label{fig:broken_add_weight}
     \vspace{-0.6cm}
\end{figure}

%% file: figures/inspection.tex
\begin{figure}[t!]
    \centering
    \includegraphics[width=0.45\textwidth]{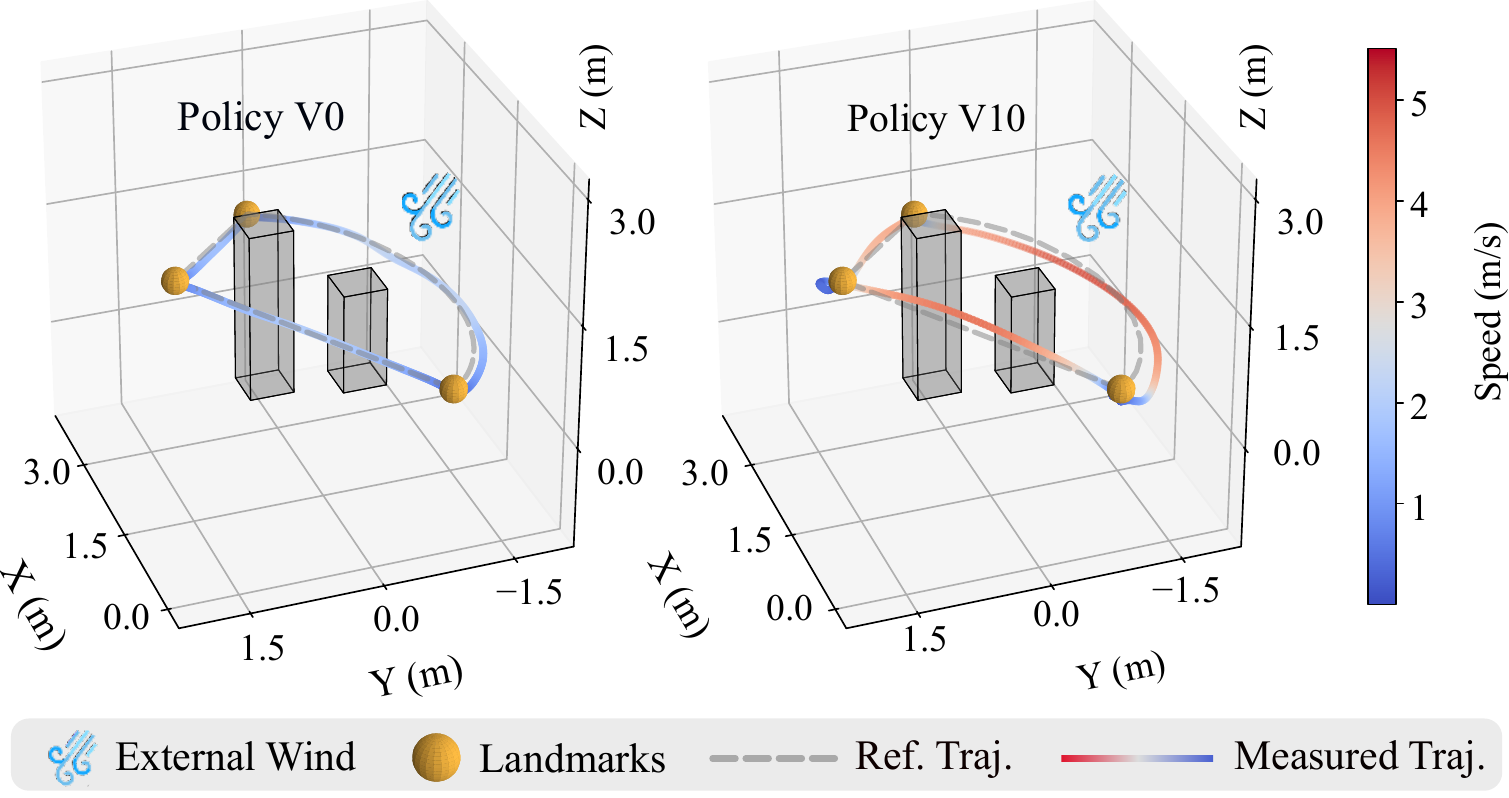}
    \caption{Evolution of flight agility during the inspection mission. The drone must traverse three landmarks while battling airflow from a fan (top-right). Starting from a baseline of \SI{12}{\second} (Policy V0, left), the framework adapts to the disturbance and accelerates the execution to \SI{7}{\second} (Policy V10, right).}
    \label{fig:inspection}
\end{figure}

%% file: figures/ablation.tex
\begin{figure}[t!]
    \centering
    \includegraphics[width=0.48\textwidth]{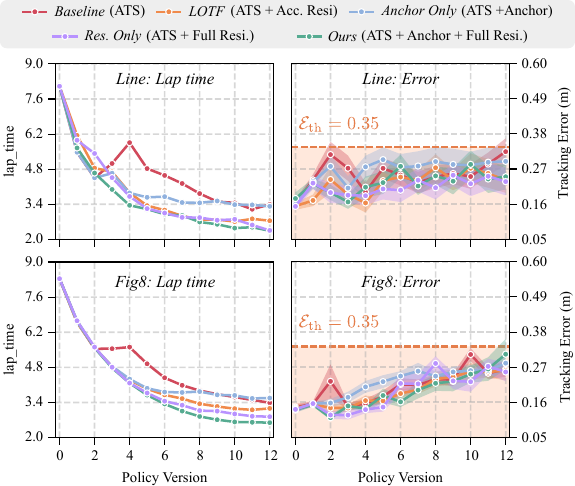}
    \caption{\textbf{Ablation study of residual dynamics and anchored rollouts.}
        \textit{Note:} Since no existing methods support online learning at the physical limit in real-world settings, all baselines are augmented with our ATS framework to enable feasible deployment.
        We compare the \textit{Baseline} (nominal), \textit{LOTF}~\cite{pan2026learning} (acceleration residual only), component ablations (\textit{Anchor/Residual Only}), and \textit{Ours} (full system).
        Enabled by our ATS framework, all methods successfully maintain a bounded tracking error less than $\mathcal E_{\text{th}}$ while reducing lap time, demonstrating the framework's effectiveness. Furthermore, \textit{Ours} achieves the shortest lap time, demonstrating that anchored initialization combined with full hybrid residual modeling is critical for pushing the agility envelope.}
    \label{fig:ablations}
    \vspace{-0.7cm}
\end{figure}

%% file: figures/fig8_wind.tex
\begin{figure*}
    \centering
    \includegraphics[width=0.95\textwidth]{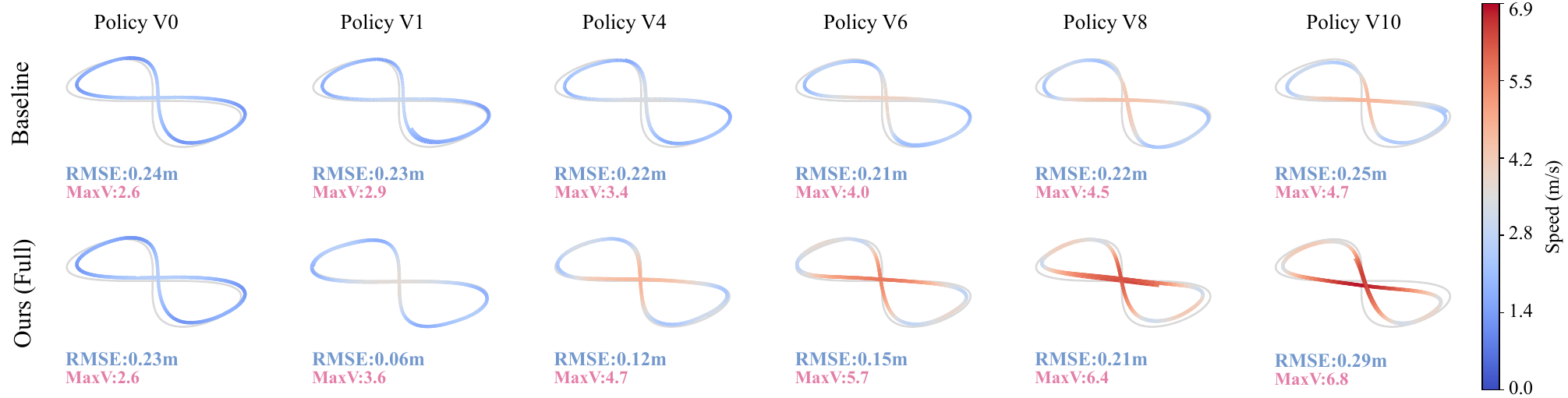}
    \caption{{\textbf{Online adaptation under wind disturbance.} 
    Comparison between the \textit{Baseline} and \textit{Ours}.
    The \textit{Baseline} fails to accelerate due to unmodeled wind drag, remaining stuck at lower speeds of $4.7$ m/s. 
    In contrast, \textit{Ours} demonstrates a rapid adaptation cycle: the residual network first suppresses the tracking error (V0 $\to$ V1, RMSE drops to \SI{0.06}{\meter}), creating a safety margin. 
    Subsequently, ATS exploits this margin to aggressively scale up agility, achieving a top speed of \SI{6.8}{\meter\per\second} (V10) while maintaining the trajectory tracking error within the threshold.}}
    \label{fig:fig8_wind}
    \vspace{-0.6cm}
\end{figure*}

%% file: sections/discussion.tex
\section{Discussion}
\label{sec:discussion}

We present a self-adaptive framework that enables autonomous quadrotors to evolve from conservative operation to the agility ceiling of their CTBR command interface \emph{in real time}. Unlike methods requiring offline identification or extensive pre-training, our approach achieves this progression within 100 seconds of flight without any prior system identification. By tightly coupling differentiable simulation, online residual learning, and Adaptive Temporal Scaling (ATS), the system autonomously bridges the gap between conservative planning and agile execution, grounding performance dynamically in the robot's instantaneous capabilities.

\subsection{Synergy between Residual Learning and ATS}
Ablation studies reveal a critical dependency: \emph{accurate short-horizon prediction is a prerequisite for safely exploiting agility}.
Without residual learning, significant model mismatch (e.g., unmodeled drag or delays) increases real-world tracking error, causing the ATS safety barrier (Eq.~\eqref{eq:potential}) to prevent time compression.
By introducing the hybrid residual model to minimize this dynamics mismatch, we expand the feasible safety margin. ATS immediately exploits this margin by decreasing $\alpha$, pushing the system toward the CTBR command-interface ceiling (i.e., commanded body-rate saturation) rather than being bounded by modeling artifacts. This mechanism explains why our full method attains substantially higher terminal speeds: improved prediction accuracy unlocks agility without sacrificing safety.

\subsection{Adaptation versus Static Robustness}
Hardware variation experiments (Fig.~\ref{fig:broken_add_weight}) highlight the distinction between \emph{static robustness} and \emph{adaptive evolution}. Classical robust control typically ensures stability under a worst-case uncertainty set, often at the cost of nominal performance. In contrast, our framework continuously identifies and optimizes for the \emph{current} dynamics.
This is evident in the propeller damage scenario: rather than maintaining flight in an overly conservative regime, the system rapidly learns the altered dynamics and re-optimizes temporal scaling. This online specialization allows the policy to adapt to degradation without overestimating capabilities or imposing unnecessary conservatism.

\subsection{Limitations and Future Work}

The present implementation has several limitations that suggest promising directions for future work. First, the saturation observed in our experiments is imposed by the CTBR command interface, not by the rotors themselves: Table~\ref{tab:theoretical_bounds} shows residual headroom between the CTBR (\SI{2.24}{\second}) and rotor-thrust (\SI{2.02}{\second}) bounds on Figure-8. Operating at the individual-rotor-thrust level could expose this headroom, at the cost of a more challenging sim-to-real transfer and safety problem. Second, ATS currently optimizes only the temporal scaling of fixed path geometries; extending it to joint spatial--temporal optimization~\cite{ren2025safety}, in the spirit of MPCC~\cite{romero2022model}, would enable more flexible adaptation and direct incorporation of obstacle-avoidance constraints. Third, the residual model is learned passively during stable task execution, leaving boundary regions of the state--action space under-modeled; safe active exploration could enable the controller to probe high-uncertainty regions while preserving stability. Finally, this work assumes accurate, low-latency state estimation, whereas aggressive flight can induce perception noise and latency; integrating perception-aware constraints into the ATS barrier would keep optimized trajectories within the sensing limits of the platform.

%% file: sections/acknowlegement.tex
\section*{Acknowledgments}
This work was supported by the European Union’s Horizon Europe Research and Innovation Programme under grant agreement No. 101120732 (AUTOASSESS) and the European Research Council (ERC) under grant agreement No. 864042 (AGILEFLIGHT). The authors thank Ismail Geles, Yixi Cai, and Leonard Bauersfeld for discussions and internal review.

%% file: references.bib
@article{Song23Reaching,
  title={Reaching the limit in autonomous racing: Optimal control versus reinforcement learning},
  author={Song, Yunlong and Romero, Angel and M{\"u}ller, Matthias and Koltun, Vladlen and Scaramuzza, Davide},
  journal={Science Robotics},
  volume={8},
  number={82},
  pages={eadg1462},
  year={2023},
}

@article{kaufmann23champion,
  title={Champion-level drone racing using deep reinforcement learning},
  author={Kaufmann, Elia and Bauersfeld, Leonard and Loquercio, Antonio and M{\"u}ller, Matthias and Koltun, Vladlen and Scaramuzza, Davide},
  journal={Nature},
  volume={620},
  number={7976},
  pages={982--987},
  year={2023},
}

@article{hwangbo2017control,
  title={Control of a quadrotor with reinforcement learning},
  author={Hwangbo, Jemin and Sa, Inkyu and Siegwart, Roland and Hutter, Marco},
  journal={IEEE Robotics and Automation Letters},
  volume={2},
  number={4},
  pages={2096--2103},
  year={2017},
  publisher={IEEE}
}

@article{eschmann2024learning,
  title={Learning to fly in seconds},
  author={Eschmann, Jonas and Albani, Dario and Loianno, Giuseppe},
  journal={IEEE Robotics and Automation Letters},
  volume={9},
  number={7},
  pages={6336--6343},
  year={2024},
  publisher={IEEE}
}

@inproceedings{nava2022fast,
  title={Fast aquatic swimmer optimization with differentiable projective dynamics and neural network hydrodynamic models},
  author={Nava, Elvis and Zhang, John Z and Michelis, Mike Yan and Du, Tao and Ma, Pingchuan and Grewe, Benjamin F and Matusik, Wojciech and Katzschmann, Robert Kevin},
  booktitle={International Conference on Machine Learning},
  pages={16413--16427},
  year={2022},
  organization={PMLR}
}

@inproceedings{xing2024bootstrapping,
  title={Bootstrapping Reinforcement Learning with Imitation for Vision-Based Agile Flight},
  author={Xing, Jiaxu and Romero, Angel and Bauersfeld, Leonard and Scaramuzza, Davide},
  booktitle={Conference on Robot Learning},
  pages={2542--2556},
  year={2025},
  organization={PMLR}
}

@inproceedings{heiden2021disect,
  author = {Eric Heiden and Miles Macklin and Yashraj S. Narang and Dieter Fox and Animesh Garg and Fabio Ramos},
  title = {{DiSECt: A Differentiable Simulation Engine for Autonomous Robotic Cutting}},
  booktitle = {Proceedings of Robotics: Science and Systems},
  year = {2021},
  address = {Virtual},
  month = {July},
  doi = {10.15607/RSS.2021.XVII.067}
}

@inproceedings{heeg2025learning,
  title={Learning quadrotor control from visual features using differentiable simulation},
  author={Heeg, Johannes and Song, Yunlong and Scaramuzza, Davide},
  booktitle={2025 IEEE International Conference on Robotics and Automation (ICRA)},
  pages={4033--4039},
  year={2025},
  organization={IEEE}
}

@article{loquercio2021learning,
  title={Learning high-speed flight in the wild},
  author={Loquercio, Antonio and Kaufmann, Elia and Ranftl, Ren{\'e} and M{\"u}ller, Matthias and Koltun, Vladlen and Scaramuzza, Davide},
  journal={Science Robotics},
  volume={6},
  number={59},
  pages={eabg5810},
  year={2021},
  publisher={American Association for the Advancement of Science}
}

@INPROCEEDINGS{bauersfeld2021neurobem, 
    AUTHOR    = {Leonard Bauersfeld AND Elia Kaufmann AND Philipp Foehn AND Sihao Sun AND Davide Scaramuzza}, 
    TITLE     = {NeuroBEM: Hybrid Aerodynamic Quadrotor Model}, 
    BOOKTITLE = {Proceedings of Robotics: Science and Systems}, 
    YEAR      = {2021}
}

@article{faessler2017differential,
  title={Differential flatness of quadrotor dynamics subject to rotor drag for accurate tracking of high-speed trajectories},
  author={Faessler, Matthias and Franchi, Antonio and Scaramuzza, Davide},
  journal={IEEE Robotics and Automation Letters},
  volume={3},
  number={2},
  pages={620--626},
  year={2017},
  publisher={IEEE}
}

@article{foehn2022agilicious,
  title={Agilicious: Open-source and open-hardware agile quadrotor for vision-based flight},
  author={Foehn, Philipp and Kaufmann, Elia and Romero, Angel and Penicka, Robert and Sun, Sihao and Bauersfeld, Leonard and Laengle, Thomas and Cioffi, Giovanni and Song, Yunlong and Loquercio, Antonio and others},
  journal={Science Robotics},
  volume={7},
  number={67},
  pages={eabl6259},
  year={2022},
  publisher={American Association for the Advancement of Science}
}

@inproceedings{xing2024contrastive,
  title={Contrastive learning for enhancing robust scene transfer in vision-based agile flight},
  author={Xing, Jiaxu and Bauersfeld, Leonard and Song, Yunlong and Xing, Chunwei and Scaramuzza, Davide},
  booktitle={2024 IEEE International Conference on Robotics and Automation (ICRA)},
  pages={5330--5337},
  year={2024},
  organization={IEEE}
}

@inproceedings{xu2023efficient,
  title={Efficient tactile simulation with differentiability for robotic manipulation},
  author={Xu, Jie and Kim, Sangwoon and Chen, Tao and Garcia, Alberto Rodriguez and Agrawal, Pulkit and Matusik, Wojciech and Sueda, Shinjiro},
  booktitle={Conference on Robot Learning},
  pages={1488--1498},
  year={2023},
  organization={PMLR}
}

@inproceedings{xu2022accelerated,
  title={Accelerated Policy Learning with Parallel Differentiable Simulation},
  author={Xu, Jie and Makoviychuk, Viktor and Narang, Yashraj and Ramos, Fabio and Matusik, Wojciech and Garg, Animesh and Macklin, Miles},
  booktitle={International Conference on Learning Representations},
  year={2022}
}

@inproceedings{luo2024residual,
  title={Residual policy learning for perceptive quadruped control using differentiable simulation},
  author={Luo, Jing Yuan and Song, Yunlong and Klemm, Victor and Shi, Fan and Scaramuzza, Davide and Hutter, Marco},
  booktitle={2025 IEEE International Conference on Robotics and Automation (ICRA)},
  pages={1--8},
  year={2025},
  organization={IEEE}
}

@inproceedings{georgiev2024adaptive,
  title={Adaptive horizon actor-critic for policy learning in contact-rich differentiable simulation},
  author={Georgiev, Ignat and Srinivasan, Krishnan and Xu, Jie and Heiden, Eric and Garg, Animesh},
  booktitle={Proceedings of the 41st International Conference on Machine Learning},
  pages={15418--15437},
  year={2024}
}

@article{tang2025deep,
  title={Deep reinforcement learning for robotics: A survey of real-world successes},
  author={Tang, Chen and Abbatematteo, Ben and Hu, Jiaheng and Chandra, Rohan and Mart{\'\i}n-Mart{\'\i}n, Roberto and Stone, Peter},
  journal={Annual Review of Control, Robotics, and Autonomous Systems},
  volume={8},
  number={1},
  pages={153--188},
  year={2025},
  publisher={Annual Reviews}
}

@inproceedings{he2024agile,
  author    = {Tairan He and Chong Zhang and Wenli Xiao and Guanqi He and Changliu Liu and Guanya Shi},
  title     = {{Agile But Safe: Learning Collision-Free High-Speed Legged Locomotion}},
  booktitle = {Proceedings of Robotics: Science and Systems},
  year      = {2024},
  month     = {July},
  address   = {Delft, Netherlands}
}

@article{chi2024universal,
  title={Universal manipulation interface: In-the-wild robot teaching without in-the-wild robots},
  author={Chi, Cheng and Xu, Zhenjia and Pan, Chuer and Cousineau, Eric and Burchfiel, Benjamin and Feng, Siyuan and Tedrake, Russ and Song, Shuran},
  journal={arXiv preprint arXiv:2402.10329},
  year={2024}
}

@article{hanover2024autonomous,
  title={Autonomous drone racing: A survey},
  author={Hanover, Drew and Loquercio, Antonio and Bauersfeld, Leonard and Romero, Angel and Penicka, Robert and Song, Yunlong and Cioffi, Giovanni and Kaufmann, Elia and Scaramuzza, Davide},
  journal={IEEE Transactions on Robotics},
  volume={40},
  pages={3044--3067},
  year={2024},
  publisher={IEEE}
}

@article{pan2026learning,
  title={Learning on the Fly: Rapid Policy Adaptation via Differentiable Simulation},
  author={Pan, Jiahe and Xing, Jiaxu and Reiter, Rudolf and Zhai, Yifan and Aljalbout, Elie and Scaramuzza, Davide},
  journal={IEEE Robotics and Automation Letters},
  year={2026},
  publisher={IEEE}
}

@inproceedings{pmlr-v305-sobanbabu25a,
  title={Sampling-based System Identification with Active Exploration for Legged Sim2Real Learning},
  author={Sobanbabu, Nikhil and He, Guanqi and He, Tairan and Yang, Yuxiang and Shi, Guanya},
  booktitle={Conference on Robot Learning},
  pages={578--598},
  year={2025},
  organization={PMLR}
}

@article{aljalbout2025reality,
  title={The Reality Gap in Robotics: Challenges, Solutions, and Best Practices},
  author={Aljalbout, Elie and Xing, Jiaxu and Romero, Angel and Akinola, Iretiayo and Garrett, Caelan Reed and Heiden, Eric and Gupta, Abhishek and Hermans, Tucker and Narang, Yashraj and Fox, Dieter and others},
  journal={Annual Review of Control, Robotics, and Autonomous Systems},
  volume={9},
  year={2025},
  publisher={Annual Reviews}
}

@article{chi2025diffusion,
  title={Diffusion policy: Visuomotor policy learning via action diffusion},
  author={Chi, Cheng and Xu, Zhenjia and Feng, Siyuan and Cousineau, Eric and Du, Yilun and Burchfiel, Benjamin and Tedrake, Russ and Song, Shuran},
  journal={The International Journal of Robotics Research},
  volume={44},
  number={10-11},
  pages={1684--1704},
  year={2025},
  publisher={Sage Publications Sage UK: London, England}
}

@inproceedings{tobin2017domain,
  title={Domain randomization for transferring deep neural networks from simulation to the real world},
  author={Tobin, Josh and Fong, Rachel and Ray, Alex and Schneider, Jonas and Zaremba, Wojciech and Abbeel, Pieter},
  booktitle={2017 IEEE/RSJ international conference on intelligent robots and systems (IROS)},
  pages={23--30},
  year={2017},
  organization={IEEE}
}

@article{lee2020learning,
  title={Learning quadrupedal locomotion over challenging terrain},
  author={Lee, Joonho and Hwangbo, Jemin and Wellhausen, Lorenz and Koltun, Vladlen and Hutter, Marco},
  journal={Science Robotics},
  volume={5},
  number={47},
  pages={eabc5986},
  year={2020},
  publisher={American Association for the Advancement of Science}
}

@article{miki2022learning,
  title={Learning robust perceptive locomotion for quadrupedal robots in the wild},
  author={Miki, Takahiro and Lee, Joonho and Hwangbo, Jemin and Wellhausen, Lorenz and Koltun, Vladlen and Hutter, Marco},
  journal={Science Robotics},
  volume={7},
  number={62},
  pages={eabk2822},
  year={2022},
  publisher={American Association for the Advancement of Science}
}

@article{ren2025safety,
  title={Safety-assured high-speed navigation for MAVs},
  author={Ren, Yunfan and Zhu, Fangcheng and Lu, Guozheng and Cai, Yixi and Yin, Longji and Kong, Fanze and Lin, Jiarong and Chen, Nan and Zhang, Fu},
  journal={Science Robotics},
  volume={10},
  number={98},
  pages={eado6187},
  year={2025},
  publisher={American Association for the Advancement of Science}
}

@article{o2022neural,
  title={Neural-fly enables rapid learning for agile flight in strong winds},
  author={O’Connell, Michael and Shi, Guanya and Shi, Xichen and Azizzadenesheli, Kamyar and Anandkumar, Anima and Yue, Yisong and Chung, Soon-Jo},
  journal={Science Robotics},
  volume={7},
  number={66},
  pages={eabm6597},
  year={2022},
  publisher={American Association for the Advancement of Science}
}

@inproceedings{shi2019neural,
  title={Neural lander: Stable drone landing control using learned dynamics},
  author={Shi, Guanya and Shi, Xichen and O’Connell, Michael and Yu, Rose and Azizzadenesheli, Kamyar and Anandkumar, Animashree and Yue, Yisong and Chung, Soon-Jo},
  booktitle={2019 IEEE International Conference on Robotics and Automation (ICRA)},
  pages={9784--9790},
  year={2019},
  organization={IEEE}
}

@article{janner2019trust,
  title={When to trust your model: Model-based policy optimization},
  author={Janner, Michael and Fu, Justin and Zhang, Marvin and Levine, Sergey},
  journal={Advances in neural information processing systems},
  volume={32},
  year={2019}
}

@article{chen2018neural,
  title={Neural ordinary differential equations},
  author={Chen, Ricky TQ and Rubanova, Yulia and Bettencourt, Jesse and Duvenaud, David K},
  journal={Advances in neural information processing systems},
  volume={31},
  year={2018}
}

@misc{jax2018github,
  author = {James Bradbury and Roy Frostig and Peter Hawkins and Matthew James Johnson and Chris Leary and Dougal Maclaurin and George Necula and Adam Paszke and Jake Vander{P}las and Skye Wanderman-{M}ilne and Qiao Zhang},
  title = {{JAX}: composable transformations of {P}ython+{N}um{P}y programs},
  url = {http://github.com/jax-ml/jax},
  version = {0.3.13},
  year = {2018},
}

@article{romero2022model,
  title={Model predictive contouring control for time-optimal quadrotor flight},
  author={Romero, Angel and Sun, Sihao and Foehn, Philipp and Scaramuzza, Davide},
  journal={IEEE Transactions on Robotics},
  volume={38},
  number={6},
  pages={3340--3356},
  year={2022},
  publisher={IEEE}
}

@article{nan2025efficient,
  title={Efficient Model-Based Reinforcement Learning for Robot Control via Online Learning},
  author={Nan, Fang and Ma, Hao and Guan, Qinghua and Hughes, Josie and Muehlebach, Michael and Hutter, Marco},
  journal={arXiv preprint arXiv:2510.18518},
  year={2025}
}

@inproceedings{buesing2018woulda,
  title={Woulda, Coulda, Shoulda: Counterfactually-Guided Policy Search},
  author={Buesing, Lars and Weber, Theophane and Zwols, Yori and Heess, Nicolas and Racaniere, Sebastien and Guez, Arthur and Lespiau, Jean-Baptiste},
  booktitle={International Conference on Learning Representations},
year = {2018}
}

@article{qu2026pragmatist,
  title={A pragmatist robot: Learning to plan tasks by experiencing the real world},
  author={Qu, Kaixian and Lan, Guowei and Zurbr{\"u}gg, Ren{\'e} and Chen, Changan and Mower, Christopher E and Bou-Ammar, Haitham and Hutter, Marco},
  journal={IEEE Robotics and Automation Letters},
  year={2026},
  publisher={IEEE}
}
